\documentclass{article}

\usepackage[T1]{fontenc}
\usepackage{lmodern}
\usepackage[verbose=true,letterpaper]{geometry}
\AtBeginDocument{
	\newgeometry{
		textheight=9in,
		textwidth=5.5in,
		top=1in,
		headheight=12pt, 
		headsep=25pt,
		footskip=30pt
	}
}

\usepackage{authblk}

\usepackage[utf8]{inputenc} % allow utf-8 input
\usepackage[T1]{fontenc}    % use 8-bit T1 fonts
\usepackage{hyperref}       % hyperlinks
\usepackage{url}            % simple URL typesetting
\usepackage{booktabs}       % professional-quality tables
\usepackage{amsfonts}       % blackboard math symbols
\usepackage{nicefrac}       % compact symbols for 1/2, etc.
\usepackage{microtype}      % microtypography

\usepackage[linesnumbered,algoruled,boxed,lined]{algorithm2e}
\usepackage{algorithmic}
\usepackage{bm}
\usepackage{amsmath}
\usepackage{amssymb}
\usepackage{amsthm}
\usepackage{graphicx}

\newcommand{\xbf}{{\mathbf x}}
\newcommand{\R}{{\mathbb{R}}}
\newcommand{\Omegabf}{{\bm{\Omega}}}
\newcommand{\omegabf}{{\bm{\omega}}}
\DeclareMathOperator*{\argmin}{\mathrm{argmin}}
\newcommand{\loss}{\mathcal{L}}
\newtheorem{theorem}{Theorem}

\newcommand{\esp}[1]{
    \underset{#1}{\mathbb{E}}\
}

\newcommand{\espdevant}[1]{
    {\mathbb{E}}_{\,#1}\
}
\title{Learning Landmark-Based Ensembles with Random Fourier Features and Gradient Boosting}

\author[1]{\hspace*{1.1cm}Léo Gautheron}
\author[2]{Pascal Germain}
\author[1]{Amaury Habrard}
\author[1]{\newline Emilie Morvant}
\author[1]{Marc Sebban}
\author[ ]{Valentina Zantedeschi}
\affil[1]{Univ Lyon, UJM-Saint-Etienne, CNRS, Institut d Optique Graduate School,\newline Laboratoire Hubert Curien UMR 5516, Saint-Etienne, France}
\affil[2]{Equipe-projet Modal, Inria Lille - Nord Europe, Villeneuve d’Ascq, France}
\begin{document}
    
    \maketitle
    
    \begin{abstract}
        We propose a Gradient Boosting algorithm for learning an ensemble of kernel functions adapted to the task at hand. 
        Unlike state-of-the-art Multiple Kernel Learning techniques that make use of a pre-computed dictionary of kernel functions to select from, at each iteration we fit a kernel by approximating it as a weighted sum of Random Fourier Features (RFF) and by optimizing their barycenter.
        This allows us to obtain a more versatile method, easier to set-up and likely to have better performance.
        Our study builds on a recent result showing one can learn a kernel from RFF by computing the minimum of a PAC-Bayesian bound on the kernel alignment generalization loss, which is obtained efficiently from a closed-form solution.
        %By studying the generalization capabilities of the so-found kernels using the PAC-Bayes theory, we also derive a closed-form formulation of a posterior distribution over the drawn RFF that is guaranteed to minimize the generalization gap.
        % We empirically validate the proposed method by showing how it outperforms both Boosting-base and kernel-learning baselines on several datasets. 
        We conduct an experimental analysis to highlight the advantages of our method w.r.t.\  both Boosting-based and kernel-learning state-of-the-art methods.
    \end{abstract}
    
    \section{Introduction}
    \label{sec:intro}
        
    Kernel methods are among the most popular approaches in machine learning due to their capability to address non-linear problems, their robustness and their simplicity. 
    However, they exhibit two main flaws in terms of memory usage and time complexity.
    To overcome the latter, some numerical approximation methods have been developed \cite{williams2001using,DBLP:journals/jmlr/DrineasM05}. 
    Landmark-based approaches \cite{BalcanBS08,BalcanBS08_COLT,bellet2012similarity,zantedeschi2018fast} can be used to reduce the number of instances to consider in order to reduce the number of comparisons~\cite{steinwart2003sparseness}, but they heavily depend on the choice of the kernel. Tuning the kernel is, however, difficult and represents another drawback to tackle. 
    Multiple Kernel Learning (MKL)~\cite{lanckriet2004learning,gonen2011multiple,xia2013mkboost,wu2017boosting} and Matching Pursuit (MP) methods~\cite{mallat1993matching,vincent2002kernel} can provide alternatives to this problem but these require the use of a pre-defined dictionary of base functions.
    
    %or the incorporation of random Fourier Features (RFF) to approximate the kernel. However, they still
    Another strategy to improve the scalability of kernel methods is to use the Random Fourier Feature (RFF) approach that proposes to approximate some invariant-shift kernel with random features based on the Fourier Transform of the kernel \cite{rahimi2008random}.
    This approach is data independent and then a predictor can be learned over these random features. 
    Several works have extended this approach by allowing one to adapt the approximation with respect to the (learning) data points \cite{yang2015carte,oliva2016bayesian,sinha2016learning,LetarteMG19,agrawal19a}.
    Among them, the recent work of
    \cite{LetarteMG19} presents a method to quickly obtain a weighting distribution over the random features by a single pass over them. 
    The method is derived from a statistical learning analysis, starting from the observation that each random feature can be interpreted as a weak hypothesis in the form of trigonometric functions obtained by the Fourier decomposition. 
    Thus, a predictor can be seen as a weighted majority vote over the random features. 
    This Fourier decomposition is then considered as a \emph{prior} distribution over the space of weak hypotheses/random features; the authors propose to learn a \emph{posterior} distribution by optimizing a PAC-Bayesian bound with respect to a kernel alignment generalization loss over the learning data points. 
    In other words, this corresponds to learning automatically a representation of the data through the approximation which then does not require to choose or tune a kernel in advance. 
    
    However, in practice, the method of \cite{LetarteMG19} requires the use of a fixed set of landmarks selected beforehand and independently from the learning task. 
    It is only once these landmarks are selected that the method can learn a representation based on the PAC-Bayesian bound.  
    This leads to three important drawbacks: {\it (i)} the need for a heuristic strategy for selecting enough relevant landmarks, {\it (ii)} these landmarks and the associated representation might not be adapted for the task at hand, and {\it (iii)} the number of landmarks might not be minimal, inducing higher computational and memory costs. 
     Instead of deliberately fixing the landmarks beforehand, we propose in this paper a Gradient Boosting approach (GB)~\cite{friedman2001greedy} for learning both the landmarks and the associated random features combination directly, leading to a strong predictor. 
     This strategy allows us to provide more compact and efficient representations in the context where the learning budget might be limited.
     % A revoir
     
     The reminder of the paper is organized as follows.
     Section~\ref{sec:notations} introduces the notations and the setting of the paper. Then, we recall in Section~\ref{sec:posterior} the work of~\cite{LetarteMG19}. 
     We introduce our landmark-based gradient boosting approach in Section~\ref{sec:our_method}. 
     The experiments are performed in Section~\ref{sec:expe}.
    Then we conclude in Section~\ref{sec:conclu}

    \section{Notations and Setting}
    \label{sec:notations}
    % In this paper, we consider boosted classifiers $H:\R^d\rightarrow Y$ with $Y=\{-1,1\}$ that are linear combinations of $T$ weak learners:
    % \begin{equation} \label{eq:comb}
    %     H(\xbf)=\text{sign}\left(\sum_{t=1}^T \alpha_t h_t(\xbf)\right)
    % \end{equation}
    % with $w_t$ the weight of the weak learner $h_t(\xbf)$ learned at the iteration $t$ from the training set $S=\{z_i=(\xbf_i,y_i)\}_{i=1}^n\sim \mathcal{D}^n$, where $\mathcal{D}$ is the unknown distribution over $\R^d\times Y$.
    % A common choice of weak learners in boosting methods \cite{schapire1999improved,friedman2001greedy} are decision trees~\cite{chen2016xgboost,ke2017lightgbm}, but few works also utilize kernel functions~\cite{xia2013mkboost,wu2017boosting}.
    We consider here binary classification tasks from a \mbox{$d$-dimensional} input space $\R^d$ to the label set $Y=\{-1,1\}$.
    Let $S=\{z_i=(\xbf_i,y_i)\}_{i=1}^n\sim \mathcal{D}^n$ be a training set of $n$ points sampled {\it i.i.d.} from $\mathcal{D}$, a fixed and unknown data-generating distribution over $\R^d\times Y$. 

In this paper, we focus on 
    %EM : manque une transition ici. Pourquoi on passe du "setting" de base aux algos avec noyaux, je sais pas mais ça me gêne
    %Standard 
    kernel-based algorithms that rely on pre-defined kernel functions $k:\R^d\times \R^d \rightarrow [-1,1]$ assessing the similarity between any two points of the input space. 
    These methods present good performances when the parameters of the kernels are learned and the chosen kernel can fit the distribution of the data. 
    % For example, for the RBF kernel $k (\xbf,\xbf')=\exp\big(-\gamma\|\xbf{-}\xbf'\|^2\big)\,,$ the parameter $\gamma$ must be carefully tuned.
    % for the polynomial kernel $k(\xbf,\xbf')=(\gamma\xbf^T\xbf'+r)^d$ the parameters $\gamma$, $r$ and $d$ must also be carefully tuned. 
    However, selecting the right kernel and tuning its parameters is computationally expensive.
    %   when the range of the values of the parameters are large as it is often required to evaluate all the possible kernels on the training set
    For this reason, Multiple Kernel Learning techniques~\cite{lanckriet2004learning,gonen2011multiple,xia2013mkboost,wu2017boosting} have been proposed to select the combination of kernels that fits best the training data:
    a dictionary of base functions $\{k_t\}_{t=1}^T$ is composed by considering various kernels with their parameters fixed to several and different values and a combination is learned, taking the following form:
    \begin{equation} \label{eq:comb}
        H(\xbf, \xbf')\ =\ \sum_{t=1}^T \alpha_t k_t(\xbf, \xbf'),
    \end{equation} 
    with $\alpha_t \in \R$ the weight of the kernel $k_t(\xbf, \xbf')$.
    
    Similarly, in our method, we aim at learning linear combinations of kernels.
    However, we do not rely on a pre-computed dictionary of kernel functions. 
    We rather learn them greedily, one per iteration of the Gradient Boosting procedure we propose (described in Section~\ref{sec:our_method}).
    % say why
    Because of the computational advantages described in Section~\ref{sec:intro}, we consider landmark-based shift-invariant kernels relying on the value $\delta=\xbf_t - \xbf \in R^d$ and denoted by abuse of notation: %of the form:
    \begin{equation} \label{eq:land-kernel}
        k(\delta) = k(\xbf_t - \xbf) = k(\xbf_t, \xbf),
    \end{equation}
    where $\xbf_t \in \R^d$ is the landmark of the input space which all the instances are compared to, that strongly characterizes the kernel.
    At each iteration of our Gradient Boosting procedure, we optimize not only this landmark but also the kernel function itself, exploiting the flexibility of the framework provided by~\cite{LetarteMG19}.
    We write the kernel as a sum of Random Fourier Features~\cite{rahimi2008random} and we learn a posterior distribution over them.
    We achieve this by studying the generalization capabilities of the so-defined functions through the lens of the PAC-Bayesian theory.
    This theoretical analysis ultimately allows us to derive a closed-form solution of the posterior distribution $q_t$ (over the RFF at a given iteration $t$), which is guaranteed to minimize the kernel alignment loss.
    In the following section, we recall the framework of~\cite{LetarteMG19} and adapt it to our scenario.
    
    \section{Pseudo-Bayesian Kernel Learning with RFF}
    \label{sec:posterior}
    % We recall in this section, the analysis provided by \citet{LetarteMG19}.
    The kernel learning method proposed by \cite{LetarteMG19} builds on the Random Fourier Features approximations proposed in~\cite{rahimi2008random}. 
    Given a shift-invariant kernel $k(\delta)=k(\xbf-\xbf')=k(\xbf,\xbf')$, \cite{rahimi2008random} show that 
    $$k(\xbf-\xbf')=\esp{\omegabf\sim p}\cos\left(\omegabf\cdot(\xbf-\xbf')\right),$$
    with $p$ the Fourier transform of $k$ defined as
    $$p(\omegabf)=\frac{1}{(2\pi)^d}\int_{\R^d}k(\delta)\exp(-i\omegabf\cdot\delta)d\delta.$$
    This allows the kernel $k$ to be approximated in practice by drawing $K$ vectors from $p$ denoted by $\Omegabf = \{\omegabf_j\}_{j=1}^K\sim p^K$ and computing $$ k(\xbf-\xbf') \simeq \frac{1}{K}\sum_{j=1}^{K}\cos\left(\omegabf_j\cdot(\xbf-\xbf')\right).$$ 
    The larger $K$, the better the resulting approximation. 
    % As example the Fourier transform of the RBF kernel is the normal distribution with mean $0$ and variance $2\gamma$:  $p = \mathcal{N}(0,2\gamma)$.
    
    Instead of drawing RFF for approximating a known kernel, \cite{LetarteMG19} propose to learn a new one by deriving a posterior distribution $q_t$ for a given landmark point in $\{\xbf_t\}_{t=1}^T$: 
    $$k_{q_t}(\xbf_t-\xbf)=\esp{\omegabf\sim q_t} \cos\left(\omegabf\cdot(\xbf_t-\xbf)\right).$$
    % where $h_\omegabf(\xbf-\xbf')=\cos\left(\omegabf\cdot(\xbf-\xbf')\right)$ can be seen as a weak hypothesis. 
    % The distribution $q$ is learned by minimizing the expected value $\loss$ of a pairwise loss: 
    % $\loss(k_q)=\esp{z,z'\sim\mathcal{D}^2}\ell(k_q(\xbf,\xbf'))$
    % with $\ell(k_q(\xbf-\xbf'))=\frac{1-yy'k_q(\xbf-\xbf')}{2}$. 
    % As the corresponding empirical loss uses two examples from the training set, these examples are dependent. 
    % This prevents the authors from bounding the expected loss $\loss(k_q)$ using its empirical counterpart with standard results from the statistical learning theory because it would require the samples to be independently drawn from the distribution. To remedy this problem, the authors rather derive independently a bound for each sample $\xbf_i$ on the expected loss where the first sample of the pairwise loss is $\xbf_i$. 
    A distribution $q_t$ is learned by minimizing a PAC-Bayesian generalization bound on the expected value of the loss between the landmark $\xbf_t$ and any point $(\xbf,y) \sim \mathcal{D}$.
    
    Let $(\xbf_t,y_t)$ be a sample, then its expected loss $\loss^t$ and empirical loss $\widehat{\loss}^t$ are respectively defined as 
    $$\loss^t=\esp{(\xbf,y)\sim \mathcal{D}}\ell(k_{q_t}(\xbf_t-\xbf)),\quad\mbox{ and }\quad \widehat{\loss}^t=\frac{1}{n-1}\sum_{j=1,j\neq t}^n\ell(k_{q_t}(\xbf_t-\xbf_j)).$$ 

    Using the PAC-Bayesian theory, they obtain the following theorem, under the linear loss \mbox{$\ell(k_{q_t}(\xbf_t-\xbf)) = \frac12 - \frac12y_tyk_{q_t}(\xbf_t-\xbf)$}, by expressing the loss as \begin{align*}
    \loss^t(k_{q_t})\ &=\ 
    \loss^t\left( \esp{\omegabf\sim p} h_\omegabf^t\right)\\ &=\ 
    \esp{\omegabf\sim p}\loss^t(h_\omegabf^t),
    \end{align*}
    with $h_\omegabf^t(\xbf)=\cos(\omegabf\cdot(\xbf_t-\xbf))$.
    We note that the result also stands for any $[0,1]$-valued convex loss $\ell$. Indeed, by Jensen's inequality, we have
    $\loss^t(k_{q_t})=
    \loss^t( \espdevant{\omegabf\sim p} h_\omegabf^t)\leq
    \espdevant{\omegabf\sim p}\loss^t(h_\omegabf^t)$.
    \begin{theorem} [Theorem 1 from \cite{LetarteMG19}]
    \label{ref:theoletarte}
        For $s>0$, $i\in\{1,\dots,n\}$, a convex loss function $\ell:\R\times\R\to[0,1]$, and a prior distribution $p$ over $\R^d$, with probability $1-\epsilon$ over the random choice of $S\sim\mathcal{D}^n$, we have for all $q$ on $\R^d$:
        $$\loss^t(k_q)\leq
         \esp{\omegabf\sim p}\widehat\loss^t(h_\omegabf^t)+\frac{1}{s}\Bigg(\text{KL}(q\Vert p) + \frac{s^2}{2(n-1)}+\ln\frac{1}{\epsilon}\Bigg),$$
where $\text{KL}(q\Vert p)=\espdevant{\omegabf\sim p}\frac{p(\omegabf)}{q(\omegabf)}$ is the Kullback-Leibler divergence between $q$ and $p$.
    \end{theorem}

 %   When the loss $\ell$ is linear, they get $\loss^i(k_q)=\esp{\omegabf\sim p}\loss^i(h_\omegabf^i)$ with $h_\omegabf^i(\xbf)=\cos(\omegabf\cdot(\xbf_i-\xbf))$. 
    It is well known~\cite{alquier2016properties,catoni2007pac,germain2009pac}
    %As a result, they find 
    that the closed form solution minimizing the bound is the pseudo-posterior distribution $Q^t$ computed as
    \begin{equation}\label{eq:closed_form_aistats}
        Q^t_j=\frac{1}{Z_t}\exp\Bigg(-\beta\sqrt{n}\widehat{\loss}^t(h_\omegabf^t)\Bigg),
    \end{equation}
    for $j=1,\dots,K$ with $\beta$ a parameter and $Z_t$ the normalization constant. Finally, given a sample point $(\xbf_t,y_t)$ and $K$ vectors $\omegabf$ denoted by $\Omegabf^t = \{\omegabf_j^t\}_{j=1}^K\sim p^K$, their kernel is finally defined as: $$k_{Q^t}(\xbf_t-\xbf)=\sum_{j=1}^{K}Q^t_j\cos(\omegabf_j\cdot(\xbf_t-\xbf)).$$
    
    Then \cite{LetarteMG19} learn a representation of the input space of $n_L$ features where each new feature $t=1,\ldots,n_L$ is computed using the kernel $k_{Q^t}$ with the sample $(\xbf_t,y_t)$. 
    To do so, they consider a set of $n_L$ landmarks $L=\{(\xbf_t,y_t)\}_{t=1}^{n_L}$ which they chose either as a random subset of the training set, or as the centers of a clustering of the training set. 
    Then, during a second step, a (linear) predictor can be learned from the new representation.

It is worth noticing that this kind of procedure exhibits two drawbacks. First, the model can  be optimized only after having learned the representation. Second, the set of landmarks $L$ has to be fixed before learning the representation.
Thus, the constructed representation is not guaranteed to be relevant for the learning algorithm considered.
To tackle these issues, we propose 
    %Then, rather than learning the new representation in a first step, and then learning a predictor in a second step, we present 
    in the next section a method performing the two steps at the same time through a gradient boosting algorithm, that allows us to 
    %and additionally, instead of defining entirely the set of landmarks $L$ before learning the representation, we further propose
     learn the set of landmarks.
  %  at each iteration the landmark that appear to be currently the best. Let us now continue with the presentation of our gradient boosting based method.
    
    \section{Gradient Boosting for Random Fourier Features}
    %\section{Proposed Method} % un titre plus sexy serait sympa ?
    % pas trop sexy 'our contribution'
    \label{sec:our_method}
    
    The approach we propose to follow the widely used gradient boosting framework first proposed by~\cite{friedman2001greedy}. 
Before presenting our contribution, we quickly recall the classical gradient boosting algorithm instantiated with the least square loss.
    
    \begin{algorithm}[t]
        \caption{Gradient Boosting with least square loss~\cite{friedman2001greedy}\label{algo:gboost}}
     \SetKwInOut{Input}{Input}\SetKwInOut{Output}{Output}
        \Input{Training set $S=\{\xbf_i,y_i\}_{i=1}^n$ with $y_i\in\{-1,1\}$;\\ $T$: number of iterations; $v$: learning rate}
        \Output{Weighted sum of predictors: $H(\xbf)=\text{sign}\left(H_0(\xbf)+\underset{t=1}{\overset{T}{\sum}}v\alpha_t h_{a_t}(\xbf)\right)$}
        \begin{algorithmic}[1]
            \STATE $\displaystyle H_0(\xbf) = \frac1n \sum_{i=1}^n y_i $
            \FOR {$t=1,\dots,T$}
            \STATE $\forall i=1,\dots,n,\quad \tilde{y}_i = y_i - H_{t-1}(\xbf_i)$
            \STATE $\displaystyle  (\alpha_t,a_t)=\argmin_{\alpha,a} \sum_{i=1}^n\big(\tilde{y}_i-\alpha h_a(\xbf)\big)^2,$ where $a$ denotes the parameters of the model $h_a$
            \STATE $H_t(\xbf) = H_{t-1}(\xbf) + v\alpha_t h_{a_t}(\xbf)$
            \ENDFOR
        \end{algorithmic}
    \end{algorithm}

    \subsection{Gradient Boosting in a Nutshell}

    Gradient boosting is an ensemble method that aims at learning a weighted majority vote over an ensemble of  predictors in a greedy way by learning iteratively the predictors to add to the ensemble.
    The final majority vote is of the form
    $$
    \forall \xbf\in\R^d,\ H(\xbf) = \text{sign}\left(H_0(\xbf)+\underset{t=1}{\overset{T}{\sum}}v\alpha_t h_{a_t}(\xbf)\right),
    $$
    where $H_0$ is a predictor fixed before the iterative process and is usually set such that it returns the same value for every data point,  and $v\alpha_t$ is the weight associated to the predictor $h_{a_t}$ ($v$ is called the learning rate\footnote{The parameter $v$ is often referred as learning rate or shrinkage parameter. 
    Decreasing $v$ usually improves the empirical performances~\cite{buhlmann2007boosting} but requires to increase the number of boosting iterations $T$.} and is fixed for each iteration, and $\alpha_t$ is called the optimal step size learned at the same time as the parameters $a_t$ of the predictor $h_{\alpha_t}$).
    %To do so, gradient boosting follows a greedy approach that learns iteratively the predictors to add to the ensemble.
    Given a differentiable loss, the objective of the gradient boosting algorithm is to perform a gradient descent where the variable to be optimized is the ensemble and the function to be minimized is the empirical loss.

    \iffalse
    The ensemble is initialized to a set that contains a single predictor $H_0(x)$ outputting the same prediction $\alpha$ for every input example. 
    Then at each iteration, a differentiable loss function $\ell$ is evaluated between the target values and the current prediction, and a predictor is trained to output the negative gradient of the loss with respect to the current prediction. 
    In other words, it performs a gradient descent in functional space where the variable to be optimized is the ensemble and the function to be minimized is the empirical loss.
    \fi
    
    We now remind the gradient boosting algorithm instantiated with the least square loss in Algorithm~\ref{algo:gboost} proposed by~\cite{friedman2001greedy}. 
    At the beginning (line {\bf 1}), the ensemble is constituted by only one predictor, the one that outputs the mean label over  the whole training set.
At each iteration, the first step (line {\bf 3}) consists in computing the negative gradient of the loss, also called the residual and denoted by $\tilde{y}_i$, for each training example $(\xbf_i,y_i)\in S$. Note that in the case of the least square loss the residual of an example is the deviation between its true label and the returned value of the current model. 
    Then, it learns the parameters $a_t$ of the predictor $h_{\alpha_t}$, along with the optimal step size $\alpha_t$, that fit the best the residuals (line {\bf 4}).
    Finally, the current model is updated by adding  $v\alpha_th_{a_t}(\cdot)$ (line {\bf 5}) to the vote.

    \iffalse
    The ensemble is initialized to output the mean label over all the training set (line {\bf 1}). 
    At each iteration the negative gradient of the loss---also called residual---$\tilde{y}_i$ is computed in line {\bf 3}. 
    Let $h_a : \R^d\rightarrow \R$ be some predictor with $a$ its parameters. 
    Then in line {\bf 4}, we learn simultaneously the optimal parameters $a_t$ of $h$ and the optimal step size $\alpha_t$ for $\alpha h_a$ to fit at best the residuals. 
    Finally, the current model is updated in line {\bf 5}. 
    In addition to the step size $\alpha_t$, the predictor $h_a$ is multiplied by a parameter $v\in(0,1]$ before being added to the ensemble. 
    This parameter is often known as learning rate or shrinkage parameter and decreasing $v$ usually improves empirical performances~\cite{buhlmann2007boosting} but requires to increase the number of boosting iterations $T$.
    \fi
    
\subsection{Our Algorithm}

We now propose to benefit from the gradient boosting to tackle the drawbacks of the landmark-based approach of~\cite{LetarteMG19} recalled in~Section~\ref{sec:posterior}. 
Our objective is to learn at the same time the landmarks ({\it i.e.}, the representation) and the classification model.

    Let $k$ be a shift-invariant kernel and let $p$ be its Fourier transform.
    At each iteration $t$, given $\Omegabf^t = \{\omegabf_j^t\}_{j=1}^K\sim p^K$ a set of $K$ random features drawn from $p$, the objective is twofold: % (line {\bf 4} of Algorithm~\ref{algo:gboost}):
    \begin{itemize}
        \item  Learn the parameters $a_t$ of the base learner $h_{a_t}$ defined as
    \begin{equation}\label{eq:predictor}
        h_{a_t}(\xbf)=\sum_{j=1}^K Q_j^t\cos\Big(\omegabf_j^t\cdot(\xbf_t-\xbf)\Big).
    \end{equation}
    In our case, the parameters to be learned are $a_t=(\xbf_t,Q^t)$ where $\xbf_t$ is a landmark, and $Q^t$ is the pseudo-posterior distribution that can be computed using a closed-form similar to Equation~\eqref{eq:closed_form_aistats}.
%In our case, the parameters to be learned are $a_t=(\Omegabf^t,\xbf_t,Q^t)$ with $\Omegabf^t = \{\omegabf_j^t\}_{j=1}^K\sim p^K$ a set of $K$ random features drawn from $p$, and $\xbf_t$ a landmark to be learned, and where $Q^t$ is the pseudo-posterior distribution to be computed using a closed-form similar to Equation~\eqref{eq:closed_form_aistats}.
\item Compute the optimal step size $\alpha_t$. 
 \end{itemize}  
In order to benefit from the theoretical guarantees of Theorem~\ref{ref:theoletarte}, and of the closed form of Equation~\eqref{eq:closed_form_aistats}, we propose the following greedy approach consisting in computing the landmark $\xbf_t$ by fixing the weight of each random features to $\frac{1}{K}$ (Equation~\eqref{eq:loss_landmark}), then $Q^t$ thanks to its closed-form (Equation~\eqref{eq:closed_form_Q}), and finally $\alpha_t$ (Equation~\eqref{eq:alpha}).

 %To do so, we propose to use the following greedy approach.
%To learn a predictor at iteration $t$ (line {\bf 4} of Algorithm \ref{algo:gboost}), we adopt a greedy approach by first learning the parameters $a_t$ of our predictor $h$ and then computing the optimal step size $\alpha_t$.
First, given the set of random features $\Omegabf^t$, we look for the landmark $\xbf_t \in \R^d$ that minimizes the average least square loss between the residuals and the kernel approximation defined as:
    \begin{equation}\label{eq:loss_landmark}
    f_{\Omegabf^t} 
    (\xbf_t)=\frac{1}{n}\sum_{i=1}^{n}\Bigg(\tilde{y_i}-\frac{1}{K}\sum_{j=1}^K \cos\Big(\omegabf_j^t\cdot(\xbf_t-\xbf_i)\Big)\Bigg)^2.
    \end{equation}
    The minimization is done by performing a gradient descent of $f_{\Omegabf^t}$ to find the landmark $\xbf_t$ that minimizes $f_{\Omegabf^t}$ where the gradient of $f_{\Omegabf^t}$ with respect to $\xbf_t$ is given by:
    \begin{equation}\label{eq:gradient_wrt_landmark}
    \frac{\partial f_{\Omegabf^t}}{\partial \xbf_t}=\frac{2}{n}\sum_{i=1}^{n}\Bigg(\frac1K\sum_{j=1}^{K}\omegabf_j^t\sin\Big(\omegabf_j^t\cdot(\xbf_t-\xbf_i)\Big)\Bigg)\Bigg(\tilde{y_i}-\frac1K\sum_{j=1}^{K}\cos\Big(\omegabf_j^t\cdot(\xbf_t-\xbf_i)\Big)\Bigg).
    \end{equation}
%It is important to notice that unlike the base learner $h_{a_t}$, the function $f_{\Omegabf^t}$ considers the kernel approximation without the weights given by $Q^t$ but where each random feature for $j=1,\dots,K$ has a uniform weight of $\frac{1}{K}$. 
 %   This is because we make the choice of not computing $Q^t$ at the same time as $\xbf_t$ during the gradient descent as we want to benefit from its closed form solution and theoretical guarantees.
    
   Second, given the landmark $\xbf_t$ found during the gradient descent, and given the set $\Omegabf^t$, we compute the pseudo-posterior distribution $Q^t$ as:
   \begin{align}\label{eq:closed_form_Q}
   \nonumber Q_j^t\ &=\ \frac{1}{Z_t}
    \exp\Big(-c f_{\omegabf_j^t}(\xbf_t) \Big)\\
    &= \ \frac{1}{Z_t} \exp\left(\frac{c}{n}\sum_{i=1}^{n}\Bigg(\tilde{y_i}- \cos\Big(\omegabf_j^t\cdot(\xbf_t-\xbf_i)\Big)\Bigg)^2\right),
    \end{align}
    for $j=1,\ldots,K$ with $c\geq 0$ a parameter and $Z_t$ the normalization constant.
% for the elements of $Q^t$ to sum to $1$.

    To finish, the optimal step size $\alpha_t$ is obtained by setting to $0$ the derivative of line {\bf 4} with respect to $\alpha$.
We then have
\begin{equation}\label{eq:alpha}
        \alpha_t\ = \ \displaystyle \frac{\displaystyle \sum_{i=1}^n\tilde{y}_ih_{a_t}(\xbf_i)}{\displaystyle \sum_{i=1}^nh_{a_t}(\xbf_i)^2}.
    \end{equation}

    This approach has two clear advantages compared to the two-step method of  \cite{LetarteMG19}, where one learns the mapping first---for a pre-defined set of landmarks---and learns the predictor afterwards.
    \begin{enumerate}
        \item Gradient Boosting allows constructing iteratively the mapping by optimizing one landmark at each step.%---the one that appears to currently be the best---at each step.
        \item The final predictor is learned at the same time and the learning procedure can be stopped when the empirical loss stops decreasing.
    \end{enumerate}
    Consequently, the final mapping is likely to be less redundant and more suitable for the task at hand.

    \section{Experiments}
    \label{sec:expe}
    
    %phrase pas claire : We conduct experiments to compare our new landmark-based learning method with RFF and gradient boosting, refered as \underline{GBRFF}, with both boosting-based methods both with the landmark learning method using RFF.
    
    In this section, we provide an empirical study of our method, referred as \underline{GBRFF}.
    Firstly, we compare the performances of \underline{GBRFF} with the two-step procedure from~\cite{LetarteMG19}, referred as~\underline{PBRFF}, and also with boosting-based methods described in the next paragraph.
    Then, we compare the influence of the number of landmarks between \underline{GBRFF} and \underline{PBRFF}.
    
    \paragraph{Experimental Setup.}
      For \underline{GBRFF}, we consider predictors as described in Equation~\eqref{eq:predictor} and select by cross-validation the parameter $c\in\{0\}\cup 2^{\{0,\dots,10\}}$.
     
    We compare \underline{GBRFF} with the following algorithms :
\begin{itemize}  
\item \underline{PBRFF}~\cite{LetarteMG19}  consists in first learning the new representation and then learning a linear SVM on the mapped training set.
    We select by cross-validation its parameters $\beta\in10^{\{-3,\dots,3\}}$ and $C\in10^{\{-3,\dots,3\}}$.
    
    \item \underline{XGB} for Xgboost ~\cite{chen2016xgboost} and \underline{LGBM} for LightGBM~\cite{ke2017lightgbm} which are state-of-the-art gradient boosting methods using trees as base predictors. For these methods, we select by cross-validation the maximum depth of the trees in $\{1,\dots,5\}$.
    
    \item \underline{MKBOOST}~\cite{xia2013mkboost} which is a Multiple Kernel Learning method based on the AdaBoost algorithm. At each boosting iteration, it selects the best performing kernel plugged inside a SVM, according to the Boosting weight distribution over the training examples.
% to classify the training examples where more weight is given to the example incorrectly classified at the previous iteration. 
As it is done by the authors, we consider at each iteration 14 RBF kernels $k(\xbf,\xbf')=\exp(-\gamma\Vert\xbf-\xbf'\Vert^2)$ with $\gamma\in2^{\{-6,\dots,7\}}$ and 3 polynomial kernels $k(\xbf,\xbf')=(\xbf^T\xbf')^d$ with $d\in\{1,2,3\}$. We select by cross-validation the SVM parameter $C\in 10^{\{-5,\dots,3\}}$.
    
    \item \underline{BMKR}~\cite{wu2017boosting} which is another Multiple Kernel Learning method based on gradient boosting with least square loss. Similarly as in \underline{MKBOOST}, it selects at each iteration the best performing kernel plugged inside an SVR to learn the residuals. It considers at each iteration 10 RBF kernels with $\gamma\in2^{\{-4,\dots,5\}}$ and the linear kernel $k(\xbf,\xbf')= \xbf^T\xbf'$. We select by cross-validation the SVR parameter $C\in 10^{\{-5,\dots,3\}}$.
    \end{itemize}
    
    For the four methods based on gradient boosting, we further select by cross-validation the learning rate $v\in\{1, 0.5, 0.1, 0.05, 0.01\}$. 
    The five boosting-based methods are run for $T=200$ iterations. 
    As for \underline{PBRFF} which is not an iterative method, we select randomly with replacement $n_L=200$ landmarks from the training set.
    For the two methods \underline{PBRFF} and \underline{GBRFF} using random features, we fix the number of random features $K$ to $100$ and we draw them from the Fourier transform of the Gaussian kernel which is the normal law.
    
    We consider $14$ datasets coming mainly from the UCI repository.
    As we deal with binary classification problems, we have binarized the datasets as described in Table \ref{tab:datasets} where the classes considered respectively as the label `-1' and as the label `+1' are 
specified. %given respectively in the columns ``Label -1" and ``Label +1".
    \begin{table}[h]\centering
        \caption{\label{tab:datasets} Description of the datasets (n: number of examples, d: number of features, c: number of classes) and the classes chosen as negative (Label -1) and positive (Label +1)}
        \resizebox{0.9\textwidth}{!}{\begin{tabular}{lrrrrrlrrrrr}
            \toprule
            Name         &     n  &  d &  c & Label -1  & Label +1 & Name         &     n  &  d &  c & Label -1 & Label +1 \\
            \midrule
            wine         &    178 & 13 &  3 &      2, 3 &        1 & wdbc         &    569 & 30 &  2 &        B &        M \\
            sonar        &    208 & 60 &  2 &         M &        R & balance      &    625 &  4 &  3 &     B, R &        L \\
            glass        &    214 & 11 &  6 & 2 3 5 6 7 &        1 & australian   &    690 & 14 &  2 &        0 &        1 \\
            newthyroid   &    215 &  5 &  3 &         1 &     2, 3 & pima         &    768 &  8 &  2 &        0 &        1 \\
            heart        &    270 & 13 &  2 &         1 &        2 & german       &   1000 & 23 &  2 &        1 &        2 \\
            bupa         &    345 &  6 &  2 &         2 &        1 & splice       &   3175 & 60 &  2 &       +1 &       -1 \\
            iono         &    351 & 34 &  2 &         g &        b & spambase     &   4597 & 57 &  2 &        0 &        1 \\
            \bottomrule
        \end{tabular}}
    \end{table}
    All datasets are normalized such that each feature has a mean of $0$ and a variance of $1$. 
    For each dataset, we generate $20$ random splits of $30\%$ training examples and $70\%$ testing examples. The hyper-parameters of all the methods are tuned by a \mbox{$5$-fold} cross-validation on the training set. 
    We report in Table~\ref{tab:results1} for each dataset the mean results over the $20$ splits. 
    In terms of accuracy, our method \underline{GBRFF} shows competitive results with the state-of-the-art as it obtains the best performances on 5 datasets out of $14$ with the best average rank among the six methods. 
    This confirms the relevance of our algorithm.

    \begin{table}[h]\centering
    \caption{\label{tab:results1}Mean test accuracy $\pm$ standard deviation over 20 random train/test splits.}
\resizebox{0.9\textwidth}{!}{\begin{tabular}{l c c c c c c}
\toprule
Dataset     & XGB  \cite{chen2016xgboost}      & LGBM  \cite{ke2017lightgbm}      & MKBOOST \cite{xia2013mkboost}   & BMKR  \cite{wu2017boosting}     & PBRFF \cite{LetarteMG19}      & GBRFF\\
\midrule
wine        &  94.92 $\pm$  2.5&  95.72 $\pm$  2.1&  98.56 $\pm$  1.3&  \textbf{99.08} $\pm$  0.6&  97.92 $\pm$  1.2&  96.80 $\pm$  2.2\\
sonar       &  76.34 $\pm$  2.8&  76.10 $\pm$  3.2&  \textbf{77.77} $\pm$  5.8&  71.78 $\pm$  4.3&  75.82 $\pm$  4.1&  76.10 $\pm$  7.5\\
glass       &  \textbf{79.70} $\pm$  3.2&  78.33 $\pm$  4.1&  78.27 $\pm$  2.8&  77.93 $\pm$  2.9&  77.27 $\pm$  3.4&  75.70 $\pm$  3.0\\
newthyroid  &  90.79 $\pm$  2.9&  83.01 $\pm$  4.3&  91.26 $\pm$ 13.8&  94.17 $\pm$  1.4&  \textbf{95.89} $\pm$  1.6&  93.18 $\pm$  1.6\\
heart       &  79.71 $\pm$  3.3&  80.74 $\pm$  2.4&  77.67 $\pm$  2.8&  \textbf{83.15} $\pm$  2.0&  83.02 $\pm$  2.2&  82.54 $\pm$  2.2\\
bupa        &  66.10 $\pm$  1.8&  67.19 $\pm$  2.7&  58.39 $\pm$  4.0&  62.11 $\pm$  3.3&  65.48 $\pm$  2.7&  \textbf{67.58} $\pm$  3.2\\
iono        &  89.25 $\pm$  1.7&  88.64 $\pm$  2.1&  91.77 $\pm$  5.9&  92.40 $\pm$  2.7&  \textbf{93.21} $\pm$  1.9&  85.55 $\pm$  2.1\\
wdbc        &  94.60 $\pm$  1.8&  95.24 $\pm$  1.8&  95.16 $\pm$  1.7&  96.20 $\pm$  0.8&  95.99 $\pm$  1.1&  \textbf{96.40} $\pm$  1.1\\
balance     &  94.13 $\pm$  2.4&  95.02 $\pm$  2.2&  83.89 $\pm$  9.3&  93.36 $\pm$  1.2&  \textbf{96.12} $\pm$  1.4&  94.77 $\pm$  1.0\\
australian  &  85.33 $\pm$  1.2&  85.65 $\pm$  1.4&  80.46 $\pm$  3.9&  85.70 $\pm$  1.1&  85.66 $\pm$  1.2&  \textbf{85.72} $\pm$  1.3\\
pima        &  75.34 $\pm$  1.8&  74.81 $\pm$  2.0&  73.06 $\pm$  2.5&  75.02 $\pm$  1.6&  75.36 $\pm$  2.1&  \textbf{75.66} $\pm$  1.9\\
german      &  71.51 $\pm$  1.2&  71.60 $\pm$  1.4&  69.84 $\pm$  1.3&  71.18 $\pm$  2.0&  71.79 $\pm$  1.3&  \textbf{72.36} $\pm$  1.5\\
splice      &  \textbf{96.35} $\pm$  0.4&  96.26 $\pm$  0.4&  82.70 $\pm$  3.8&  86.42 $\pm$  0.6&  85.27 $\pm$  0.5&  88.16 $\pm$  0.5\\
spambase    &  94.20 $\pm$  0.3&  \textbf{94.25} $\pm$  0.3&  90.45 $\pm$  0.6&  92.34 $\pm$  0.5&  91.60 $\pm$  0.4&  92.33 $\pm$  0.3\\
\midrule
Average Rank& 3.40& 3.20& 4.40& 3.07& 2.80& 2.73\\
\bottomrule
\end{tabular}}
    \end{table}
    
\begin{figure}[h]\centering   
	\makebox[\textwidth]{\includegraphics[width=\textwidth]{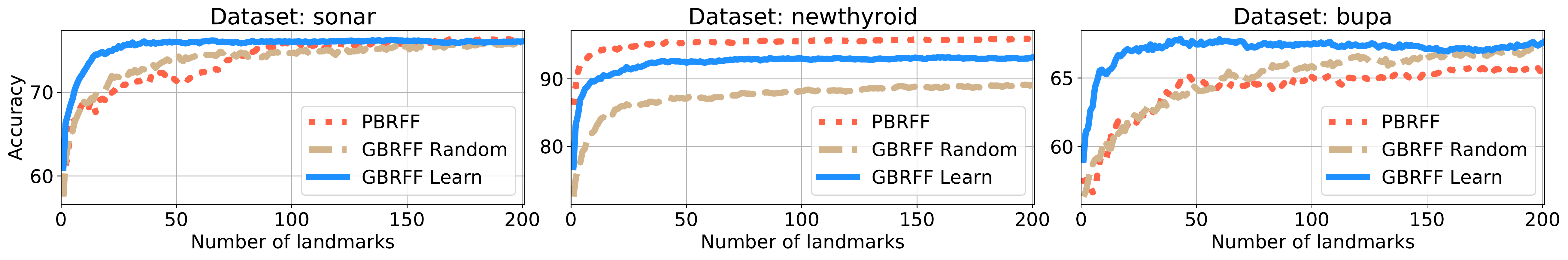}} \\   
	\caption{Mean test accuracy over $20$ train/test splits on the ``sonar'', ``newthyroid'' and ``bupa'' datasets as a function of the number of landmarks used with \underline{PBRFF} and our two variants of \underline{GBRFF}.}  \label{fig:graph_newthyroid_bupa} 
    \end{figure}

    \paragraph{Influence of the number of landmarks.}
    In Figure~\ref{fig:graph_newthyroid_bupa}, we analyze the accuracy of our landmark-based method \underline{GBRFF} in two variants. 
    The first one named \underline{GBRFF Learn} corresponds to what was done in the previous experiment where at each iteration a landmark was learned. The second named \underline{GBRFF Random} considers at each iteration a landmark drawn randomly from the training set. 
    In addition, we compare our method to \underline{PBRFF} which also draws the landmarks randomly from the training set.
    To gain relevant insights, the analysis is made on three datasets for which our method has better and worse performances compared to \underline{PBRFF}.  We consider the datasets ``sonar'', ``newthyroid'' and ``bupa''.
    
    Overall, as expected, the larger the quantity of landmarks, the better the performances for all methods. We see on the three datasets that \underline{GBRFF Learn} presents better performances than \underline{GBRFF Random}. The difference is especially large when the number of landmarks is small. For ``sonar'' and ``bupa'',  \underline{PBRFF} requires much more landmarks than \underline{GBRFF Learn} to reach its maximal value. This shows the importance of learning the landmarks compared to selecting them randomly as it allows converging faster to possibly better performances. On the other hand, the results on ``newthyroid'' are better for \underline{PBRFF}, no matter the number of landmarks used. This may happen because the linear classifier in the two methods is learned differently: it is learned using all landmarks by \underline{PBRFF} with a Linear SVM and learned one landmark at a time by \underline{GBRFF} with gradient boosting. %Depending on the task at hand, different methods can be more beneficial.
    
      \begin{figure}[h]\centering   
    	\makebox[\textwidth]{\includegraphics[width=\textwidth]{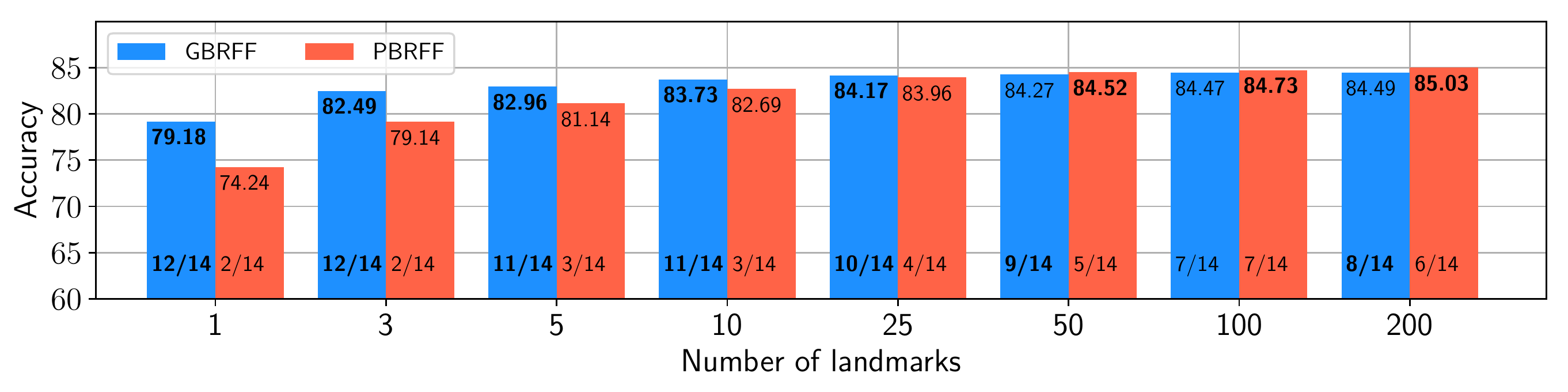}} \\   
    	\caption{Mean test accuracy over $20$ train/test splits and over the $14$ datasets as a function of the number of landmarks used to train the two methods \underline{PBRFF} and  \underline{GBRFF}. The mean values are displayed at the top of the bars, and the numbers of datasets where a method has the best performances are displayed at the bottom of the bars}  \label{fig:bars_increase_number_landmarks} 
    \end{figure}
    We summarize in Figure~\ref{fig:bars_increase_number_landmarks} the influence of the number of landmarks used to train  \underline{PBRFF} and  \underline{GBRFF}. 
    The figure gives the mean test accuracy across all datasets and over the $20$ train/test splits. 
    As seen in the previous experiment, with $200$ landmarks, \underline{PBRFF} and   \underline{GBRFF} have similar performances with respectively $84.49\%$ and $85.03\%$ of mean accuracy and with better performances for \underline{GBRFF} on 8 datasets out of $1$4. 
    However, when the number of landmarks decreases,  \underline{GBRFF} demonstrates a clear superiority. Indeed, we can observe in Figure~\ref{fig:bars_increase_number_landmarks} that with $2$5 landmarks, \underline{GBRFF} provides a mean accuracy $0.21$ point higher than the one of \underline{PBRFF} while being better on $1$0 datasets, with $10$ landmarks it is $1.04$ points higher and better for $11$ datasets, with $5$ landmarks it gets $1.82$ points higher and is still better for $11$ datasets, with $3$ landmarks it is $3.35$ points higher and finally with only one landmark it is superior with $4.94$ points higher. Additionally, our method obtains the best performances in $12$ datasets out of $14$ with less than $3$ landmarks. 
    Thus, the smaller the number of landmarks used, the better our method \underline{GBRFF} compared to \underline{PBRFF}. 
    The gain is significative when the number of landmarks is smaller than 25 which also corresponds to learning very small representations. This shows the clear advantage of our landmark-based method when learning compact representations with few landmarks, especially when one has a limited budget.
    In this case, learning the landmarks to solve the task at hand is preferable to selecting them randomly.
    
    %%% caption pour table avant table
    %% caption pour figure après figure

    \section{Conclusion}
    \label{sec:conclu}
    In this paper, we propose a Gradient Boosting algorithm where a kernel is learned at each iteration; the kernel being expressed with random Fourier features (RFF).
    Compared to state-of-the-art Multiple Kernel Learning techniques that select the best kernel function from a dictionary, and then plug it inside a kernel machine, we directly consider a kernel as a predictor that outputs a similarity to a point called landmark.
    We learn at each iteration a landmark by approximating the kernel as a sum of Random Fourier Features to fit the residuals of the gradient boosting procedure.
    Building on a recent work~\cite{LetarteMG19}, we  learn a pseudo-distribution over the RFF through a closed-form solution that minimizes a PAC-Bayes bound to induce a new kernel function tailored for the task at hand.
    The experimental study shows the competitiveness of the proposed method with state-of-the-art boosting and kernel learning methods, especially when the number of iterations used to train our model is small. 
    
    So far, the landmarks have been learned without any constraint. 
    A promising future line of research is to add a regularization on the set of landmarks to foster diversity.
    In addition, the optimization of a landmark at each iteration can be computationally expensive when the number of iterations is large, and a possibility to speed-up the learning procedure is to derive other kernel approximations where the landmarks can be computed with a closed-form solution.
    Other possibilities regarding the scalability include the use of standard gradient boosting tricks \cite{ke2017lightgbm,chen2016xgboost} such as sampling or learning the kernels in parallel.
    Another perspective could be to extend the analysis of~\cite{LetarteMG19} along with our algorithm to random Fourier features for operator-valued kernels~\cite{brault2016random} useful for multi-task learning or structured output.

   % TROP SIMILAIRE A AISTATS!!!
   % An important research direction is to derive guarantees to the final predictor, which could in turn be the bedrock of a new one-step learning procedure (in the vein of~\citet{yang2015carte,oliva2016bayesian}).
  %  Other research directions include the study of the RKHS associated with the learned kernel, and the extension of our study to wavelet transforms\citep{Mallat-book}.
    %Furthermore, considering the Fourier transform of a kernel as a \mbox{(pseudo-)}Bayesian prior might lead to other original contributions.
    %Among them, we foresee new perspectives on representation and metric learning, namely for unsupervised learning.
   % Finally, we believe that learning random feature-based representation along with ``good'' landmarks for a given task can be helpful for other (semi-)supervised classification tasks, such as learning with imbalanced data or transfer learning.

    % perspectives: 
    %derive guarantees on the overall GB, use standard GB tricks for accelerating it, add regularization on the set of landmarks to foster diversity 
    % stopping criteria ?
    
    \subsubsection*{Acknowledgments}
    This work was supported in part by the French Project
    APRIORI ANR-18-CE23-0015.
    
    \small
    
    \bibliographystyle{plain}
    \bibliography{biblio}
    
\end{document}